\documentclass[runningheads]{llncs}
\usepackage[T1]{fontenc}
\usepackage{cite}
\usepackage{graphicx}
\usepackage{multirow}
\usepackage[dvipsnames]{xcolor}
\usepackage{hyperref}

\hypersetup{hypertex=true,
            colorlinks=true,
            linkcolor=blue,
            anchorcolor=blue,
            citecolor=blue}
\usepackage[misc]{ifsym}
\usepackage{fontawesome}
\usepackage{color}
\usepackage{amsfonts}
\begin{document}
\title{Privacy-preserving Early Detection of Epileptic Seizures in Videos}
\titlerunning{Privacy-preserving Early Detection of Epileptic Seizures in Videos}
%

\author{Deval, Mehta\inst{1,2,3}\textsuperscript{(\faEnvelopeO)}
\and Shobi Sivathamboo\inst{4,5,6} \and Hugh Simpson\inst{4,5} \and Patrick Kwan\inst{4,5,6} \and Terence O'Brien\inst{4,5} \and Zongyuan Ge\inst{1,2,3,7}}
%
\authorrunning{D. Mehta et al.}
%
\institute{AIM for Health Lab, Faculty of IT, Monash University, Melbourne, Australia \url{https://www.monash.edu/it/aimh-lab/home} \\ \email{deval.mehta@monash.edu} \and  Monash Medical AI, Monash University, Melbourne, Australia \and Faculty of Engineering, Monash University, Melbourne, Australia \\ \and 
Department of Neuroscience, Central Clinical School, Faculty of Medicine Nursing and Health Sciences, Monash University, Melbourne, Australia \\ \and
Department of Neurology, Alfred Health, Melbourne, Australia \\ \and Departments of Medicine and Neurology, The University of Melbourne, Royal Melbourne Hospital, Parkville, Victoria, Australia \\ \and Airdoc-Monash Research Lab, Monash University, Melbourne, Australia}
%
\maketitle              
\begin{abstract}
In this work, we contribute towards the development of video-based epileptic seizure classification by introducing a novel framework (SETR-PKD), which could achieve privacy-preserved early detection of seizures in videos. Specifically, our framework has two significant components - (1) It is built upon optical flow features extracted from the video of a seizure, which encodes the seizure motion semiotics while preserving the privacy of the patient; (2) It utilizes a transformer based progressive knowledge distillation, where the knowledge is gradually distilled from networks trained on a longer portion of video samples to the ones which will operate on shorter portions. Thus, our proposed framework addresses the limitations of the current approaches which compromise the privacy of the patients by directly operating on the RGB video of a seizure as well as impede real-time detection of a seizure by utilizing the full video sample to make a prediction. Our SETR-PKD framework could detect tonic-clonic seizures (TCSs) in a privacy-preserving manner with an accuracy of \textbf{83.9\%} while they are only \textbf{half-way} into their progression. Our data and code is available at \url{https://github.com/DevD1092/seizure-detection}.

\end{abstract}
\section{Introduction}
Epilepsy is a chronic neurological condition that affects more than 60 million people worldwide in which patients experience epileptic seizures due to abnormal brain activity~\cite{moshe2015epilepsy}. Different types of seizures are associated with the specific part of the brain involved in the abnormal activity~\cite{fisher2017operational}. Thus, accurate detection of the type of epileptic seizure is essential to epilepsy diagnosis, prognosis, drug selection and treatment. Concurrently, real-time seizure alerts are also essential for caregivers to prevent potential complications, such as related injuries and accidents, that may result from seizures. Particularly, patients suffering from tonic-clonic seizures (TCSs) are at a high risk of sudden unexpected death in epilepsy (SUDEP)~\cite{nashef2012unifying}. Studies have shown that SUDEP is caused by severe alteration of cardiac activity actuated by TCS, leading to immediate death or cardiac arrest within minutes after the seizure~\cite{devinsky2016sudden}. Therefore, it is critical to accurately and promptly detect and classify epileptic seizures to provide better patient care and prevent any potentially catastrophic events.

The current gold standard practice for detection and classification of epileptic seizures is the hospital-based Video EEG Monitoring (VEM) units~\cite{shih2018indications}. However, this approach is expensive and time consuming which is only available at specialized centers~\cite{cascino2002video}. To address this issue, the research community has developed automated methods to detect and classify seizures based on several modalities - EEG~\cite{fan2018detecting,yuan2018multi}, accelerometer~\cite{kusmakar2018automated}, and even functional neuroimaging modalities such as fMRI~\cite{rashid2020use} and electrocorticography (ECoG)~\cite{siddiqui2019novel}. Although, there have been developments of approaches for the above modalities, seizure detection using videos remains highly desirable as it involves no contact with the patient and is easier to setup and acquire data compared to other modalities. Thus, researchers have also developed automated approaches for the video modality.

Initial works primarily employed hand-crafted features based on patient motion trajectory by attaching infrared reflective markers to specific body key points~\cite{cunha2016neurokinect,karayiannis2006automated}. However, these approaches were limited in performance due to their inability to generalize to changing luminance (night time seizures) or when the patient is occluded (covered by a bed sheet)~\cite{kalitzin2012automatic}. Thus, very recently deep learning (DL) models have been explored for this task~\cite{ahmedt2018hierarchical,ahmedt2018deep,yang2021video,perez2021transfer,hou2022automated}. ~\cite{yang2021video} demonstrated that DL models could detect generalized tonic-clonic seizures (GTCSs) from the RGB video of seizures. Authors in~\cite{perez2021transfer} radically used transfer learning (from action recognition task) to train DL networks for distinguishing focal onset seizures (FOSs) from bilateral TCSs using features extracted from the RGB video of seizures. Whereas, the authors in~\cite{hou2022automated} developed a DL model to discriminate dystonia and emotion in videos of Hyperkinetic seizures. However, these developed approaches have two crucial limitations - (1) As these approaches directly operate on RGB videos, there is a possibility of privacy leakage of the sensitive patient data from videos. Moreover, obtaining consent from patients to share their raw RGB video data for building inter-cohort validation studies and generalizing these approaches on a large scale becomes challenging; (2) The current approaches consider the full video of a seizure to make predictions, which makes early detection of seizures impossible. The duration of a seizure varies significantly among patients, with some lasting as short as 30 seconds while others can take minutes to self-terminate. Thus, it is unrealistic to wait until the completion of a long seizure to make a prediction and alert caregivers.

In this work, we address the above two challenges by building an in-house dataset of privacy-preserved extracted features from a video and propose a framework for early detection of seizures. Specifically, we investigate two aspects - (1) The feasibility of detecting and classifying seizures based only on \emph{optical flow}, a modality that captures temporal differences in a scene while being intrinsically privacy-preserving. (2) The potential of predicting the type of seizure during its progression by analyzing only a fraction of the video sample. Our early detection approach is inspired by recent developments in early action recognition in videos~\cite{furnari2020rulstm,perez2021transfer,zheng2023egocentric,guan2023egocentric,wang2021oadtr,osman2021slowfast}. We develop a custom feature extractor-transformer framework, named \textbf{SE}izure \textbf{TR}ansformer (SETR) block for processing a single video sample. To achieve early detection from a fraction of the sample, we propose \textbf{P}rogressive \textbf{K}nowledge \textbf{D}istillation (PKD), where we gradually distill knowledge from SETR blocks trained on longer portions of a video sample to SETR blocks which will operate on shorter portions. We evaluate our proposed SETR-PKD framework on two datasets - an in-house dataset collected from a VEM unit in a hospital and a publicly available dataset of video-extracted features (GESTURES)~\cite{perez2021transfer}. Our experiments demonstrate that our proposed SETR-PKD framework can detect TCS seizures with an accuracy of \textbf{83.9\%} in a privacy-preserving manner when they are only \textbf{half-way} into their progression. Furthermore, we comprehensively compare the performance of direct knowledge distillation with our PKD approach on both optical flow features (in-house dataset) and raw video features (public dataset). We firmly believe that our proposed method makes the first step towards developing a privacy-preserving real-time system for seizure detection in clinical practice.

\section{Proposed Method}
In this section, we first outline the process of extracting privacy-preserving information from RGB video samples to build our in-house dataset. Later, we explain our proposed approach for early detection of seizures in a sample.

\subsection{Privacy Preserving Optical Flow Acquisition}
Our in-house dataset of RGB videos of patients experiencing seizures resides on hospital premises and is not exportable due to the hospital's ethics agreement\footnote{We have a data ethics agreement approved for collection of data at hospital}. To work around this limitation, we develop a pipeline to extract optical flow information~\cite{horn1981determining} from the videos. This pipeline runs locally within the hospital and preserves the privacy of the patients while providing us with motion semiotics of the seizures. An example of the extracted optical flow video sample can be seen in Fig~\ref{fig1}. We use the TV-L1 algorithm~\cite{perez2013tv} to extract the optical flow features for each video, which we then export out of the hospital for building our proposed approach. We provide more information about our dataset, including the number of patients and seizures, annotation protocol, etc. in section~\ref{sec3}.

\subsection{Early Detection of Seizures in a Sample}
Consider an input optical flow video sample $V_{i}$ as shown in Fig~\ref{fig1}(a) with a time period of $T_{i}$, consisting of $N$ frames - $\{f_{0}, f_{1}, ... f_{N-1}\}$, and having a ground truth label of $y_{i}$ $\in$ $\{0, 1, ... C\}$ where is $C$ the total number of categories. Then, the task of early detection is to build a framework that could classify the category of the sample correctly by analyzing the least possible partial segment of the sample. Thus, to define the problem of early detection, we split the sample $V_{i}$ into $k$ segments -$\{0,1,...k-1\}$ starting from the beginning to the end as shown in Fig~\ref{fig1}(b). Here $V^{k-1}_{i}$ corresponds to the full video sample and the descending segments correspond to the reduced partial video samples. We build these partial segments by equally adding the temporal information throughout the sample i.e. the time period for a partial subset $V^{j}_{i}$ of a sample $V_{i}$ is computed as $(j+1) \times T_{i}/k$. Thus, the early detection task is to correctly predict the category $y_{i}$ of the sample $V_{i}$ from the lowest possible ($j$) partial segment $V^{j}_{i}$ of $V_{i}$. In Fig~\ref{fig1}, we illustrate our proposed framework where - (a) First, we build a Seizure Transformer (SETR) block for processing a single optical flow video sample (b) Later, we employ SETR based Progressive Knowledge Distillation (SETR-PKD) to achieve early detection in a sample.

\subsubsection{Processing a Single Sample}
\begin{figure}[t!]
\centering
\includegraphics[width=0.925\textwidth]{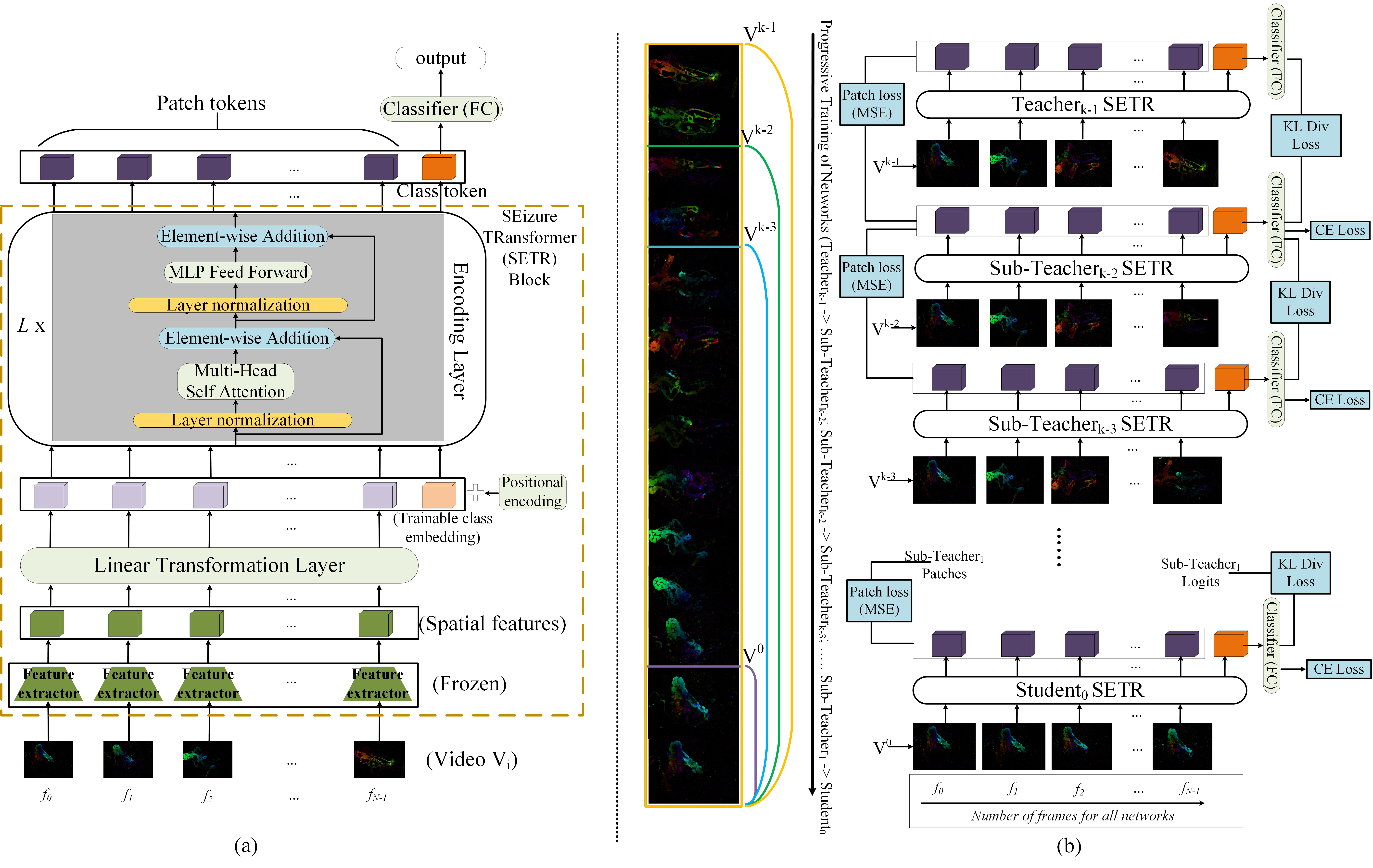}
\caption{Our proposed framework - (a) SEizure TRansformer (SETR) block for a single optical flow video sample (b) SETR based Progressive Knowledge Distillation (SETR-PKD) for early detection of seizures in a sample. (Best viewed in zoom and color).} \label{fig1}
\end{figure}

Since seizure patterns comprise of body movements, we implement transfer learning from a feature extractor pre-trained on action recognition task to extract the spatial features from the optical flow frames. Prior work~\cite{perez2021transfer} has shown that Temporal Segment Networks (TSNs)~\cite{wang2018temporal} pretrained on RGB videos of various actions are effective at extracting features from videos of seizures. We also utilize TSNs but pretrained on the optical flow modality, since we have privacy-preserved optical flow frames. The TSNs extract a 1D feature sequence for each frame $f_{j}$, referred as spatial features in Fig~\ref{fig1}(a). The spatial features are then processed by a linear transformation (1-layer MLP) that maps them into $motion_{tokens}$ $\in$ $\mathbb{R}^{N \times D}$, where each token has $D$-dimensions.

We leverage transformers to effectively learn temporal relations between the extracted spatial features of the seizure patterns. Following the strategy of ViT~\cite{dosovitskiy2020image}, after extracting the spatial features, we append a trainable class embedding $class_{embed}$ $\in$ $\mathbb{R}^{D}$ to the motion tokens. This class embedding serves to represent the temporal relationships between the motion tokens and is later used for classification ($class_{token}$ in Fig~\ref{fig1}(a)). As the order of the $motion_{tokens}$ is not known, we also add a learnable positional encoding $L_{POS}$ $\in$ $\mathbb{R}^{(N+1)\times D}$ to the combined $motion_{tokens}$ and $class_{embed}$. This is achieved using an element-wise addition and we term it as the input $X_{i}$ for the input sample $V_{i}$.

To enable the interaction between tokens and learn temporal relationships for input sample classification, we employ the Vanilla Multi-Head Self Attention (MHSA) mechanism~\cite{vaswani2017attention}. First, we normalize the input sequence $X_{i}$ $\in$ $\mathbb{R}^{(N+1)\times D}$ by passing it through a layer normalization, yielding $X^{'}_{i}$. We then use projection matrices $(Q_{i},K_{i},V_{i})$ = $(X^{'}_{i}W^{Q}_{i},X^{'}_{i}W^{K}_{i},X^{'}_{i}W^{V}_{i})$ to project $X^{'}_{i}$ into queries (Q), keys (K), and values (V), where $W^{Q/K/V}_{i}$ $\in$ $\mathbb{R}^{D\times D}$ are the projection matrices for query, key, and value respectively. Next, we compute a dot product of $Q$ with $K$ and apply a softmax layer to obtain weights on the values. We repeat this self-attention computation $N_{h}$ times, where $N_{h}$ is the number of heads, and concatenate their outputs.  Eq~\ref{eq1},~\ref{eq2} depict the MHSA process in general.

\begin{equation}\label{eq1}
A_{i} = Softmax(Q_{i}K_{i})
\end{equation}

\begin{equation}\label{eq2}
MHSA(X^{'}_{i}) = A_{i} \times W^{V}_{i}, \> \> \> \> \> \> \> \> \> \> \> \> \> \> X^{'}_{i} = Norm(X_{i})
\end{equation}

Subsequently, the output of MHSA is passed to a two-layered MLP with GELU non-linearity while applying layer normalization and residual connections concurrently. Eq~\ref{eq3},~\ref{eq4} represent this overall process.

\begin{equation}\label{eq3}
m^{'}_{l} = MHSA(X^{'}_{l-1}) + X_{l-1}, \> \> \> \> \> \> \> \> \> \> \> \> \> \> l = 1...L
\end{equation}

\begin{equation}\label{eq4}
m_{l} = MLP(Norm(m^{'}_{l})) + m^{'}_{l}, \> \> \> \> \> \> \> \> \> \> \> \> \> \> l = 1...L
\end{equation}

where $m_{L}$ $\in$ $\mathbb{R}^{(N+1)\times D}$ are the final output feature representations and $L$ is the total number of encoding layers in the Transformer Encoder. Note that the first $\mathbb{R}^{N\times D}$ features correspond to the $patch_{tokens}$, while the final $\mathbb{R}^{D}$ correspond to the $class_{token}$ of the $m_{L}$ as shown in Fig~\ref{fig1}(a). As mentioned earlier, we then use a one-layer MLP to predict the class label from the $class_{token}$. We refer to this whole process as a SEizure TRansformer (SETR) block shown in Fig~\ref{fig1}(a).

\subsubsection{Progressive Knowledge Distillation}
To achieve early detection, we use \textbf{K}nowledge \textbf{D}istillation in a \textbf{P}rogressive manner (PKD), starting from a SETR block trained on a full video sample and gradually moving to a SETR block trained on a partial video sample, as shown in Fig~\ref{fig1}(b). Directly distilling from a SETR block which has seen a significantly longer portion of the video (say $V^{k-1}_{i}$) to a SETR block which has only seen a smaller portion of the video sample (say $V^{0}_{i}$) will lead to considerable mismatches between the features extracted from the two SETRs as there is a large portion of the input sample that the $student_{0}$ SETR has not seen. In contrast, our proposed PKD operates in steps. First we pass the knowledge from teacher ($Teacher_{k-1}$ in Fig~\ref{fig1}(b)) SETR trained on $V^{k-1}_{i}$ to a student ($Sub-teacher_{k-2}$) SETR that operates on $V^{k-2}_{i}$; Later, the $Sub-teacher_{k-2}$ SETR passes its distilled knowledge to its subsequent student ($Sub-teacher_{k-3}$) SETR, and this continues until the final $Sub-teacher_{1}$ SETR passes its knowledge to the bottom most $Student_{0}$ SETR. Since the consecutive segments of the videos do not differ significantly, PKD is more effective than direct distillation, which is proven by results in section~\ref{sec3p4}.


For distilling knowledge we consider both class token and patch tokens of the teacher and student networks. A standard Kullback–Leibler divergence ($\mathcal{L}_{KL}$) loss is applied between the probabilities generated from class token of the teacher and student SETR, whereas a mean squared error ($\mathcal{L}_{MSE}$) loss is computed between the patch tokens of teacher and student SETR. Overall, a student SETR is trained with three losses - $\mathcal{L}_{KL}$ and $\mathcal{L}_{MSE}$ loss for knowledge distillation, and a cross-entropy ($\mathcal{L}_{CE}$) loss for classification, given by the equations below.

\begin{equation}
\mathcal{L}_{KL} = \tau ^ 2 \sum_{j} q^{T}_{j}(log(q^{T}_{j}/q^{S}_{j}))
\end{equation}

where $q^{S}_{j}$ and $q^{T}_{j}$ are the soft probabilities (moderated by temperature $\tau$) of the student and teacher SETRs for the $j^{th}$ class, respectively.

\begin{equation}
\mathcal{L}_{mse}= (\sum ^{N}_{i=0} |p^{T}_{i} - p^{S}_{i}\| ^{2}) / N
\end{equation}
where $N$ is the number of patches and $p^{T}_{i}$ and $p^{S}_{i}$ are the patches of teacher and student SETRs respectively.

\begin{equation}
\mathcal{L}_{total} =  \mathcal{L}_{CE} + \alpha\mathcal{L}_{KL} + \beta\mathcal{L}_{mse}  
\end{equation}

where $\alpha$ and $\beta$ are the weights for $\mathcal{L}_{KL}$ and $\mathcal{L}_{MSE}$ loss respectively.

\section{Datasets \& Experimental Results}\label{sec3}
\subsection{In-house and Public Dataset}
Our in-house dataset\footnote{We plan to release the in-house optical flow dataset and corresponding code.} contains optical flow information extracted from high-definition (1920x1080 pixels at 30 frames per second) video recordings of TCS seizures (infrared cameras are used for nighttime seizures) in a VEM unit in hospital. To annotate the dataset, two neurologists examined both the video and corresponding EEG to identify the clinical seizure onset ($t_{ON}$) and clinical seizure offset ($t_{OFF}$) times for each seizure sample. We curated a dataset comprising of 40 TCSs from 40 epileptic patients, with one sample per patient. The duration (in seconds) of the 40 TCSs in our dataset ranges from 52 to 367 s, with a median duration of 114 s. We also prepared normal samples (no seizure) for each patient by considering the pre-ictal duration from ($t_{ON}$ - 300) to ($t_{ON}$ - 60) seconds, resulting in dataset of 80 samples (40 normal and 40 TCSs). We refrain from using the 60 seconds prior to clinical onset as it corresponds to the transition period to the seizure containing preictal activity~\cite{sivathamboo2020cardiorespiratory,huberfeld2011glutamatergic}. We use a 5-fold cross validation (split based on patients) for training and testing on our dataset.

We also evaluate the effectiveness of our early detection approach on the GESTURES dataset~\cite{perez2021transfer}, which contains features extracted from RGB video samples of seizures. The dataset includes two seizure types - 106 focal onset seizures (FOS) and 77 Tonic-Clonic Seizures (TCS). In contrast to our in-house dataset, the features are provided by the authors, and we directly input them into our SETR block without using a feature extractor. To evaluate our method, we adopt the stratified 10-fold cross-validation protocol as used in GESTURES.

\subsection{Training Implementation \& Evaluation Metrics}
We implement all experiments in PyTorch 1.8.1 on a single A100 GPU. The SETR block takes in a total of 64 frames ($N$) with 512 1-D spatial feature per frame, has 8 MHSA heads ($N_{h}$) with a dropout rate of 0.1, 3 encoder layers ($L$), and 256 hidden dimensions ($D$). For early detection, we experiment by progressively segmenting a sample into -\{4,8,16\} parts ($k$). We employ a grid search to select the weight of 0.2 and 0.5 for KL divergence ($\tau$ = 10) and MSE loss respectively. We train all methods with a batch size of 16, a learning rate of 1e-3 and use the AdamW optimizer with a weight decay of 1e-4 for a total 50 epochs. For GESTURES dataset, we implement a weighted BCE loss to deal with the dataset imbalance, whereas for our in-house dataset we implement the standard BCE loss. We use precision, recall and f1-score for benchmarking.

\subsection{Performance for Early Detection}\label{sec3p3}
Table~\ref{tab1} shows the benchmarking performance of all techniques with varying fractions of input video samples on both datasets. We observed three key findings from the results in Table~\ref{tab1}. First, transformer-based methods such as our proposed \textbf{SETR-PKD} and OaDTR exhibit better performance retention compared to LSTM-based techniques (RULSTM, Slowfast RULSTM, EgoAKD, GESTURES) with a reduction in the fraction of input sample. Second, \textbf{SETR-PKD} performance increases with $k$=8 from $k$=4, but saturates at $k$=16 for in-house dataset, whereas it achieves the best performance for $k$=4 for GESTURES dataset. The median seizure length for the in-house dataset and GESTURES dataset is 114 seconds and 71 seconds, respectively. As a result,

\begin{table}[b!]
\caption{Benchmarking of different techniques for different fraction \{\textbf{1/4, 1/2, 3/4, Full}\} of input video sample. The performance is presented as mean of - \{\textbf{Precision/Recall/F1-score}\} across the 5-folds \& 10-folds for in-house and GESTURES dataset respectively. (Best viewed in zoom).}  \label{tab1}
\resizebox{\textwidth}{!}{\begin{tabular}{c|c|c|c|c|c|c|c|c}
\hline
\multirow{2}{*}{Method/Dataset}&\multicolumn{4}{c|}{In-house dataset} &\multicolumn{4}{|c}{GESTURES}\\
\cline{2-9}
 & 1/4 & 1/2 & 3/4 & Full & 1/4 & 1/2 & 3/4 & Full \\
 \hline
 RULSTM~\cite{furnari2020rulstm} & 0.57/0.56/0.56 & 0.72/0.71/0.71 & 0.79/0.79/0.79 & 0.95/0.93/0.94 & 0.65/0.64/0.64 & 0.71/0.73/0.72 & 0.84/0.85/0.84 & 0.93/0.94/0.93 \\
 Slowfast RULSTM~\cite{osman2021slowfast} & 0.57/0.56/0.56 & 0.73/0.72/0.72 & 0.81/0.80/0.80 & 0.94/0.94/0.94 & 0.67/0.65/0.66 & 0.73/0.72/0.72 & 0.86/0.84/0.85 & 0.97/0.95/0.96\\
 EgoAKD~\cite{zheng2023egocentric} & 0.64/0.65/0.64 & 0.79/0.80/0.79 & 0.89/0.90/0.89 & 0.95/0.94/0.94 & 0.70/0.69/0.69 & 0.80/0.79/0.79 & 0.93/0.90/91 & 0.97/0.94/0.95 \\
OaDTR~\cite{wang2021oadtr} & 0.66/0.65/0.65 & 0.82/0.83/0.82 & 0.90/0.90/0.90 & 0.95/\textcolor[rgb]{0,0,1}{0.95}/\textcolor[rgb]{0,0,1}{0.95} & 0.72/0.69/0.70 & 0.82/0.83/0.82 & 0.91/0.92/0.91 & \textcolor[rgb]{0,0,1}{0.99/0.99/0.99} \\
GESTURES~\cite{perez2021transfer} & 0.59/0.60/0.59 & 0.74/0.73/0.73 & 0.82/0.83/0.82 & 0.94/0.94/0.94 & 0.68/0.66/0.66 & 0.74/0.72/0.73 & 0.86/0.85/0.85 & 0.97/0.99/0.98 \\
\hline
\textbf{SETR} & 0.61/0.60/0.60 & 0.75/0.73/0.74 & 0.84/0.83/0.83 & 0.96/0.95/0.95 & 0.67/0.66/0.66 & 0.73/0.74/0.73  & 0.88/0.88/0.88 & 0.98/0.99/0.98 \\
\textbf{SETR-PKD (k=4)} & 0.63/0.62/0.62 & 0.78/0.79/0.78 & 0.89/0.90/0.89 & 0.96/0.95/0.95 & \textcolor[rgb]{0,0,1}{0.74/0.73/0.73} & \textcolor[rgb]{0,0,1}{0.86/0.85/0.85} & \textcolor[rgb]{0,0,1}{0.96/0.95/0.95} & 0.98/0.99/0.98\\
\textbf{SETR-PKD (k=8)} & \textcolor[rgb]{0,0,1}{0.70/0.69/0.69} & \textcolor[rgb]{0,0,1}{0.86/0.84/0.85} & \textcolor[rgb]{0,0,1}{0.92/0.93/0.92} & \textcolor[rgb]{0,0,1}{0.96/0.95/0.95} & 0.73/0.74/0.73 & 0.85/0.85/0.85 & 0.95/0.96/0.95 & 0.98/0.99/0.98 \\
\textbf{SETR-PKD (k=16)} & 0.69/0.69/0.69 & 0.85/0.84/0.84 & 0.92/0.92/0.92 & 0.96/0.95/0.95 & 0.72/0.73/0.72 & 0.85/0.84/0.84 & 0.96/0.95/0.95 & 0.98/0.99/0.98 \\
\hline
\end{tabular}}
\end{table}

\noindent PKD using relatively longer partial segments ($k$=4) is sufficient for GESTURES, while shorter partial segments ($k$=8) are required for our dataset. Thus, the optimal value of $k$ for PKD may vary depending on a dataset. Finally, we observed better performance on the GESTURES dataset, which is expected given the more detailed and refined features extracted from RGB video compared to optical flow information.

\subsection{Progressive v/s Direct Knowledge Distillation}\label{sec3p4}

\begin{figure}
\centering
\includegraphics[width=0.92\textwidth]{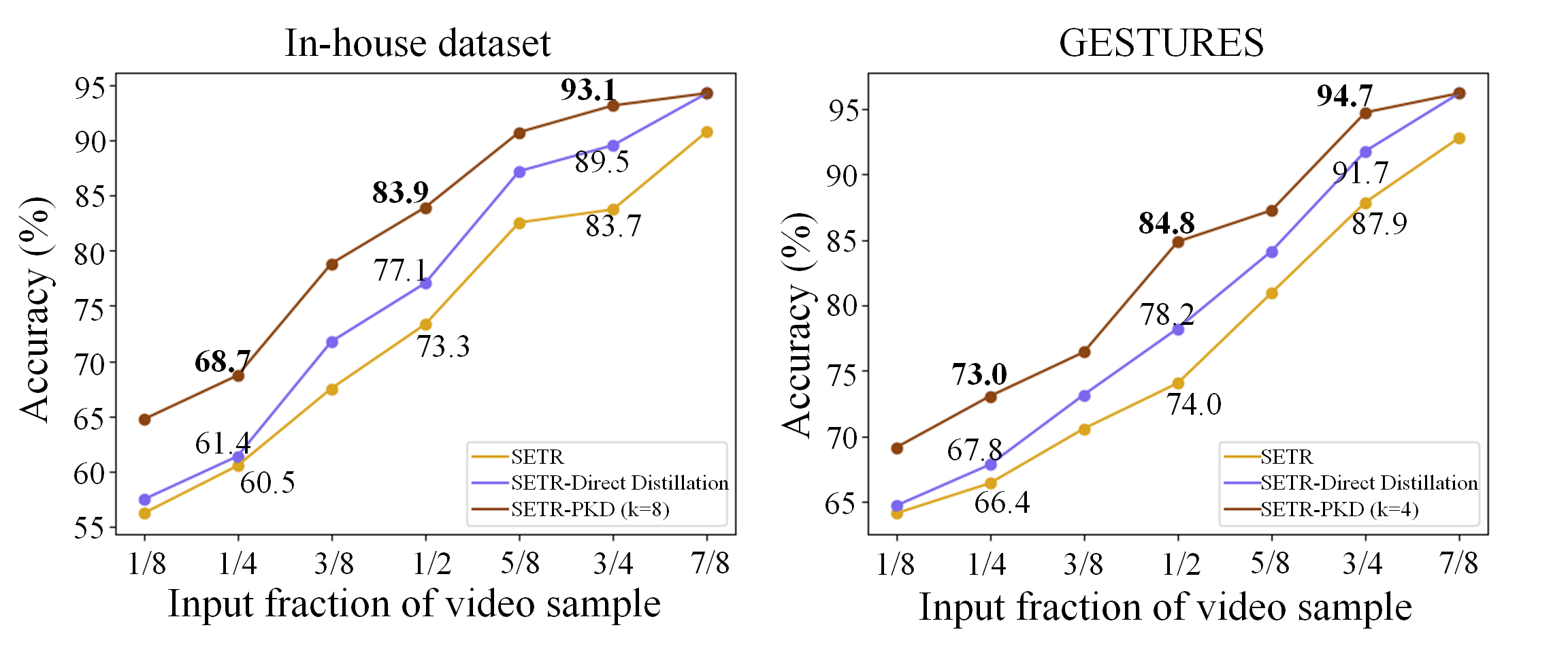}
\caption{Performance comparison of direct knowledge distillation and progressive knowledge distillation between SETR blocks for different fractions of input video sample.} \label{fig2}
\end{figure}

To validate our approach of progressive knowledge distillation in a fair manner, we conducted an ablation study to compare it with direct knowledge distillation. Fig~\ref{fig2} shows the comparison of the accuracy of the two approaches for different fractions of the input video sample on both datasets. The results indicate that although direct knowledge distillation can increase performance, it is less effective when the knowledge gap is wide, i.e., from a SETR block trained on a full input sample to a SETR block trained on a minimal fraction of the input sample (1/8, 1/4, .. 1/2) compared to when the knowledge gap is small (5/8, .. 7/8). On the other hand, our SETR-PKD approach significantly improves performance for minimal fractions of input samples on both datasets.

%



\section{Conclusion}
In this work, we show that it is possible to detect epileptic seizures from optical flow modality in a privacy-preserving manner. Moreover, to achieve real-time seizure detection, we specifically develop a novel approach using progressive knowledge distillation which proves to detect seizures more accurately during their progression itself. We believe that our proposed privacy-preserving early detection of seizures will inspire the research community to pursue real-time seizure detection in videos as well as facilitate inter-cohort studies.

%
%
%
\bibliographystyle{splncs04}
\bibliography{ref}

\begin{thebibliography}{10}
\providecommand{\url}[1]{\texttt{#1}}
\providecommand{\urlprefix}{URL }
\providecommand{\doi}[1]{https://doi.org/#1}

\bibitem{ahmedt2018hierarchical}
Ahmedt-Aristizabal, D., Fookes, C., Denman, S., Nguyen, K., Fernando, T.,
  Sridharan, S., Dionisio, S.: A hierarchical multimodal system for motion
  analysis in patients with epilepsy. Epilepsy \& Behavior  \textbf{87},
  46--58 (2018)

\bibitem{ahmedt2018deep}
Ahmedt-Aristizabal, D., Nguyen, K., Denman, S., Sridharan, S., Dionisio, S.,
  Fookes, C.: Deep motion analysis for epileptic seizure classification. In:
  2018 40th Annual International Conference of the IEEE Engineering in Medicine
  and Biology Society (EMBC). pp. 3578--3581. IEEE (2018)

\bibitem{cascino2002video}
Cascino, G.D.: Video-eeg monitoring in adults. Epilepsia  \textbf{43},  80--93
  (2002)

\bibitem{cunha2016neurokinect}
Cunha, J.P.S., Choupina, H.M.P., Rocha, A.P., Fernandes, J.M., Achilles, F.,
  Loesch, A.M., Vollmar, C., Hartl, E., Noachtar, S.: Neurokinect: a novel
  low-cost 3dvideo-eeg system for epileptic seizure motion quantification. PloS
  one  \textbf{11}(1),  e0145669 (2016)

\bibitem{devinsky2016sudden}
Devinsky, O., Hesdorffer, D.C., Thurman, D.J., Lhatoo, S., Richerson, G.:
  Sudden unexpected death in epilepsy: epidemiology, mechanisms, and
  prevention. The Lancet Neurology  \textbf{15}(10),  1075--1088 (2016)

\bibitem{dosovitskiy2020image}
Dosovitskiy, A., Beyer, L., Kolesnikov, A., Weissenborn, D., Zhai, X.,
  Unterthiner, T., Dehghani, M., Minderer, M., Heigold, G., Gelly, S., et~al.:
  An image is worth 16x16 words: Transformers for image recognition at scale.
  arXiv preprint arXiv:2010.11929  (2020)

\bibitem{fan2018detecting}
Fan, M., Chou, C.A.: Detecting abnormal pattern of epileptic seizures via
  temporal synchronization of eeg signals. IEEE Transactions on Biomedical
  Engineering  \textbf{66}(3),  601--608 (2018)

\bibitem{fisher2017operational}
Fisher, R.S., Cross, J.H., French, J.A., Higurashi, N., Hirsch, E., Jansen,
  F.E., Lagae, L., Mosh{\'e}, S.L., Peltola, J., Roulet~Perez, E., et~al.:
  Operational classification of seizure types by the international league
  against epilepsy: Position paper of the ilae commission for classification
  and terminology. Epilepsia  \textbf{58}(4),  522--530 (2017)

\bibitem{furnari2020rulstm}
Furnari, A., Farinella, G.M.: Rolling-unrolling lstms for action anticipation
  from first-person video. IEEE Transactions on Pattern Analysis and Machine
  Intelligence (PAMI)  (2020)

\bibitem{guan2023egocentric}
Guan, W., Song, X., Wang, K., Wen, H., Ni, H., Wang, Y., Chang, X.: Egocentric
  early action prediction via multimodal transformer-based dual action
  prediction. IEEE Transactions on Circuits and Systems for Video Technology
  (2023)

\bibitem{horn1981determining}
Horn, B.K., Schunck, B.G.: Determining optical flow. Artificial intelligence
  \textbf{17}(1-3),  185--203 (1981)

\bibitem{hou2022automated}
Hou, J.C., Thonnat, M., Bartolomei, F., McGonigal, A.: Automated video analysis
  of emotion and dystonia in epileptic seizures. Epilepsy Research
  \textbf{184},  106953 (2022)

\bibitem{huberfeld2011glutamatergic}
Huberfeld, G., Menendez de~la Prida, L., Pallud, J., Cohen, I., Le~Van~Quyen,
  M., Adam, C., Clemenceau, S., Baulac, M., Miles, R.: Glutamatergic pre-ictal
  discharges emerge at the transition to seizure in human epilepsy. Nature
  neuroscience  \textbf{14}(5),  627--634 (2011)

\bibitem{kalitzin2012automatic}
Kalitzin, S., Petkov, G., Velis, D., Vledder, B., da~Silva, F.L.: Automatic
  segmentation of episodes containing epileptic clonic seizures in video
  sequences. IEEE transactions on biomedical engineering  \textbf{59}(12),
  3379--3385 (2012)

\bibitem{karayiannis2006automated}
Karayiannis, N.B., Tao, G., Frost~Jr, J.D., Wise, M.S., Hrachovy, R.A.,
  Mizrahi, E.M.: Automated detection of videotaped neonatal seizures based on
  motion segmentation methods. Clinical Neurophysiology  \textbf{117}(7),
  1585--1594 (2006)

\bibitem{kusmakar2018automated}
Kusmakar, S., Karmakar, C.K., Yan, B., O’Brien, T.J., Muthuganapathy, R.,
  Palaniswami, M.: Automated detection of convulsive seizures using a wearable
  accelerometer device. IEEE Transactions on biomedical engineering
  \textbf{66}(2),  421--432 (2018)

\bibitem{moshe2015epilepsy}
Mosh{\'e}, S.L., Perucca, E., Ryvlin, P., Tomson, T.: Epilepsy: new advances.
  The Lancet  \textbf{385}(9971),  884--898 (2015)

\bibitem{nashef2012unifying}
Nashef, L., So, E.L., Ryvlin, P., Tomson, T.: Unifying the definitions of
  sudden unexpected death in epilepsy. Epilepsia  \textbf{53}(2),  227--233
  (2012)

\bibitem{osman2021slowfast}
Osman, N., Camporese, G., Coscia, P., Ballan, L.: Slowfast rolling-unrolling
  lstms for action anticipation in egocentric videos. In: Proceedings of the
  IEEE/CVF International Conference on Computer Vision. pp. 3437--3445 (2021)

\bibitem{perez2013tv}
P{\'e}rez, J.S., Meinhardt-Llopis, E., Facciolo, G.: Tv-l1 optical flow
  estimation. Image Processing On Line  \textbf{2013},  137--150 (2013)

\bibitem{perez2021transfer}
P{\'e}rez-Garc{\'\i}a, F., Scott, C., Sparks, R., Diehl, B., Ourselin, S.:
  Transfer learning of deep spatiotemporal networks to model arbitrarily long
  videos of seizures. In: Medical Image Computing and Computer Assisted
  Intervention--MICCAI 2021: 24th International Conference, Strasbourg, France,
  September 27--October 1, 2021, Proceedings, Part V 24. pp. 334--344. Springer
  (2021)

\bibitem{rashid2020use}
Rashid, M., Singh, H., Goyal, V.: The use of machine learning and deep learning
  algorithms in functional magnetic resonance imaging—a systematic review.
  Expert Systems  \textbf{37}(6),  e12644 (2020)

\bibitem{shih2018indications}
Shih, J.J., Fountain, N.B., Herman, S.T., Bagic, A., Lado, F., Arnold, S.,
  Zupanc, M.L., Riker, E., Labiner, D.M.: Indications and methodology for
  video-electroencephalographic studies in the epilepsy monitoring unit.
  Epilepsia  \textbf{59}(1),  27--36 (2018)

\bibitem{siddiqui2019novel}
Siddiqui, M.K., Islam, M.Z., Kabir, M.A.: A novel quick seizure detection and
  localization through brain data mining on ecog dataset. Neural Computing and
  Applications  \textbf{31},  5595--5608 (2019)

\bibitem{sivathamboo2020cardiorespiratory}
Sivathamboo, S., Constantino, T.N., Chen, Z., Sparks, P.B., Goldin, J.,
  Velakoulis, D., Jones, N.C., Kwan, P., Macefield, V.G., O'Brien, T.J.,
  et~al.: Cardiorespiratory and autonomic function in epileptic seizures: a
  video-eeg monitoring study. Epilepsy \& Behavior  \textbf{111},  107271
  (2020)

\bibitem{vaswani2017attention}
Vaswani, A., Shazeer, N., Parmar, N., Uszkoreit, J., Jones, L., Gomez, A.N.,
  Kaiser, {\L}., Polosukhin, I.: Attention is all you need. Advances in neural
  information processing systems  \textbf{30} (2017)

\bibitem{wang2018temporal}
Wang, L., Xiong, Y., Wang, Z., Qiao, Y., Lin, D., Tang, X., Van~Gool, L.:
  Temporal segment networks for action recognition in videos. IEEE transactions
  on pattern analysis and machine intelligence  \textbf{41}(11),  2740--2755
  (2018)

\bibitem{wang2021oadtr}
Wang, X., Zhang, S., Qing, Z., Shao, Y., Zuo, Z., Gao, C., Sang, N.: Oadtr:
  Online action detection with transformers. In: Proceedings of the IEEE/CVF
  International Conference on Computer Vision. pp. 7565--7575 (2021)

\bibitem{yang2021video}
Yang, Y., Sarkis, R.A., El~Atrache, R., Loddenkemper, T., Meisel, C.:
  Video-based detection of generalized tonic-clonic seizures using deep
  learning. IEEE Journal of Biomedical and Health Informatics  \textbf{25}(8),
  2997--3008 (2021)

\bibitem{yuan2018multi}
Yuan, Y., Xun, G., Jia, K., Zhang, A.: A multi-context learning approach for
  eeg epileptic seizure detection. BMC systems biology  \textbf{12}(6),  47--57
  (2018)

\bibitem{zheng2023egocentric}
Zheng, N., Song, X., Su, T., Liu, W., Yan, Y., Nie, L.: Egocentric early action
  prediction via adversarial knowledge distillation. ACM Transactions on
  Multimedia Computing, Communications and Applications  \textbf{19}(2),  1--21
  (2023)

\end{thebibliography}
\end{document}